\RequirePackage{amsmath}
\documentclass[sigconf]{acmart}
\usepackage{listings}
\usepackage{subfigure}
\usepackage{relsize}
\usepackage{xcolor}
\usepackage{color}
\usepackage{graphicx}
\usepackage{enumitem}
\usepackage{float}
\usepackage{xtab,booktabs}
\usepackage{soul}
\usepackage{todonotes}
\usepackage{mathtools}
\DeclarePairedDelimiter{\ceil}{\lceil}{\rceil}
\usepackage{acronym}
\usepackage[T1]{fontenc}
\usepackage[lined,commentsnumbered]{algorithm2e}
\LinesNumbered
\usepackage{algcompatible}
\SetKwFor{For}{for (}{) $\lbrace$}{$\rbrace$}

\setcopyright{rightsretained}

\acmDOI{10.1145/nnnnnnn.nnnnnnn}

\acmISBN{978-x-xxxx-xxxx-x/YY/MM}

\acmPrice{15.00}

\begin{document}

\title{The Factory Must Grow: Automation in Factorio}


\author{Kenneth N. Reid}
\orcid{0001-8654-2430}
\affiliation{%
  \institution{Michigan State University}
  \city{East Lansing} 
  \state{MI} 
  \postcode{48823}
}
\email{Ken@kenreid.co.uk}

\author{Iliya Miralavy}
\affiliation{%
  \institution{Michigan State University}
  \city{East Lansing} 
  \state{MI} 
  \postcode{48823}
}

\author{Stephen Kelly}
\affiliation{%
  \institution{Michigan State University}
  \city{East Lansing} 
  \state{MI} 
  \postcode{48823}
}

\author{Wolfgang Banzhaf}
\orcid{0002-6382-3245}
\affiliation{%
  \institution{Michigan State University}
  \city{East Lansing} 
  \state{MI} 
  \postcode{48823}
}

\author{Cedric Gondro}
\affiliation{%
  \institution{Michigan State University}
  \city{East Lansing} 
  \state{MI} 
  \postcode{48823}
}

\renewcommand{\shortauthors}{Reid, Miralavy, Kelly, Banzhaf and Gondro}


\begin{abstract}
    Efficient optimization of resources is paramount to success in many problems faced today. In the field of operational research the efficient scheduling of employees; packing of vans; routing of vehicles; logistics of airlines and transport of materials can be the difference between emission reduction or excess, profits or losses and feasibility or unworkable solutions. The video game Factorio, by Wube Software, has a myriad of problems which are analogous to such real-world problems, and is a useful simulator for developing solutions for these problems. In this paper we define the logistic transport belt problem and define mathematical integer programming model of it. We developed an interface to allow optimizers in any programming language to interact with Factorio, and we provide an initial benchmark of logistic transport belt problems. We present results for Simulated Annealing, quick Genetic Programming and Evolutionary Reinforcement Learning, three different meta-heuristic techniques to optimize this novel problem. 
\end{abstract}



\copyrightyear{2021} 
\acmYear{2021} 
\setcopyright{acmlicensed}
\acmConference{}
\acmPrice{15.00}
\acmDOI{10.1145/3321707.3321769}
\acmISBN{978-1-4503-6111-8/19/07}

\maketitle

\newacro{ARN}[ARN]{Artificial Regulatory Network}
\newacro{ASS}[ASS]{Automated Sortation Conveyor}
\newacro{D2D}[D2D]{Device to Device}
\newacro{GA}[GA]{Genetic Algorithm}
\newacro{GP}[GP]{Genetic Programming}
\newacro{HC}[HC]{Hard Constraint}
\newacro{PSO}[PSO]{Particle Swarm Optimization}
\newacro{NPC}[NPC]{Non-Player Character}
\newacro{RCON}[RCON]{Remote Console}
\newacro{SA}[SA]{Simulated Annealing}
\newacro{SC}[SC]{Soft Constraint}
\newacro{RL}[RL]{Reinforcement Learning}
\newacro{ERL}[ERL]{Evolutionary Reinforcement Learning}
\newacro{TPG}[TPG]{Tangled Program Graphs}
\newacro{FOI}[FOI]{Factorio Optimizer Interface}
\newacro{PCG}[PCG]{Procedural Content Generation}
\newacro{OR}[OR]{Operational Research}
\newacro{PSA}[PSA]{Parallel Simulated Annealing}
\newacro{qGP}[qGP]{q Genetic Programming}

\section{Introduction}\label{section:intro}

    Factorio is a video game produced by Wube software\footnote{\url{https://factorio.com/}}, which was fully released in August 2020. The premise of Factorio is an engineer who crash-lands into an alien planet, and must use their engineering expertise to accumulate resources, automate production and launch a rocket to escape the planet, all while avoiding the aliens defending their world from the pollution created by the engineers machinery.
    
    Games represent a wide variety of challenging and interesting problems designed to engage human intelligence. As such, games provide a rich test bed for Artificial Intelligence. Yannakakis and Togelius \cite{yannakakis2018} structure the application of AI in games into three categories: 1) AI for game playing; 2) AI for content generation; and 3) AI for player modelling. Of particular interest to this work are the first two categories on which we shall focus from here on.
    Of particular interest to this work are
    the first two categories on which we shall focus from here on. 
 
    In AI for \textbf{Game Playing}, artificial agents learn to interact with a game, observing the state of play and taking actions in an attempt to maximize one or more objectives. In general the objective of an artificial game-player may be to win the game (e.g. maximize game score or length of play) or to define behavior of a \ac{NPC} in the game such that the game experience is more engaging for a human player. Game playing is the oldest and likely the most common application of AI in games, and a wide array of literature exists on the subject. Some recent and well-known examples include artificial soccer players in RoboCup \cite{Asada2019}, mastering the board game Go with Artificial Neural Networks \cite{alphazero}, and playing classic Atari games at a level comparable to human players \cite{mnih15}. Artificial game playing agents have many practical applications, from discovering bugs in computer games or other interactive software \cite{chrabaszcz2018}, to autonomous robotics \cite{Dickinson2007}.
    
    \textbf{\ac{PCG}} refers to the use of AI to create game content automatically, with minimal human input. Game content may include environments, difficulty levels, rules, or characters. There are many compelling reasons to employ PCG. When game design is automated, one could imagine  building a game that never ends, with PCG continuously creating new content. PCG could also support games tailored to individual players, with the goal of improving the entertainment value and/or educational value of a game \cite{Cruz2017}. In short, PCG extends the creativity of human game designers and can therefore be used to study computational creativity more generally, see, e.g., \cite{Phillips2016}. 
    
    Factorio provides a breadth of opportunities for AI Game Playing and PCG that directly transfer to real-world problem solving scenarios. To simplify the complex mass of problems facing the player, herein mentioned problems include the maximization of the players research progress; the efficient usage of resources, space, the minimization of pollution, and more. In particular, the link between \ac{OR} and environmental issues is of a potential use of Factorio \cite{bloemhof1995interactions}. Factorio is essentially an open-ended collection of operational research and optimization problems. 
    
    Much like an ecology, Factorio can be thought of as a large set of varied interlinking problems and solutions, where the key to success is often balancing between solving these problems. For example, focusing solely on production will likely increase the pollution output, causing a game failure. 
    
    In this paper, we provide an overview of some such problems and provide an interface by which users can connect external problem solvers to Factorio. As an example of this interface we tackle a small but common set of problems: the logistic belt problems. Belts transport raw resources and manufactured goods between collection points, machinery and storage. With hundreds of items in the game which can be delivered this way,  often requiring multiple belts dedicated to each item, this is both a complex and common problem facing players. Additionally, there are different types of belts, each with different potential speeds. The purpose of these belts is to deliver goods, but also to meet specific criteria: the correct ratio of items output by the solution; materials reaching the correct end destinations; belts avoiding obstacles; and minimizing the number of belts used in a solution. 
    
    In this paper we solve several logistic belt problems of varying grid sizes and number of obstacles, by employing various metaheuristics through the interface to test the generated solutions within Factorio itself. Solving problems which are analogies for real-world problems is beneficial ...
    

\section{Related Work}\label{section:relatedWork}

   There are several works in other fields which mention Factorio: virtual economy analysis \cite{mizutani2019toward}, video game learning and community building \cite{kunze2019video}, and a consideration of the capitalistic and production-obsessed players of Factorio in \cite{leblanc2018only}. In a poster by Duhan et al \cite{duhan2019factory} the concept of creating a reinforcement learning based engine to play Factorio is described, with the potential for creation of transferable knowledge to real-world scenarios, though no further work has yet been published.
   
   In Factorio logistic belts are used to solve problems of transporting materials. A clear similarity between this contribution and real-world problems is the use of conveyor belts, their purpose, functionality and optimization. Conveyor belts, in the real world, have been referred to as productive and a particularly economical transport mechanism with high performance compared to alternatives and a small carbon footprint \cite{grinvcova2014experimental}. A recent survey paper details the importance of \ac{ASS} \cite{boysen2019automated} - highlighting their efficiency, time-savings and the reduction in employee costs, with more details on the importance of logistics for transporting items via belts. The overland conveyor is often used in transporting raw materials; such as ore before processing. Much of the literature on the optimization of such devices is with regard to design of the machine itself, and the energy consumption of fully-loaded and empty conveyors, rather than planning logistical routes \cite{gerard2009optimisation} \cite{carr2020reducing}. \ac{PSO} was also used in a similar optimization of screw conveyors, a type of conveyors used to transport granular bulk goods, but also considered design of the machinery rather than logistical placement and application \cite{cai2011optimisation}. 
   
   Routing problems are relevant to this paper as the problem being solved also requires a connection between two points on a grid. A masters thesis by one of the Wube Software developers, Ond\v{r}ej Majerech \cite{majerech2017solving}, which considers the efficient path planning of numerous agents in a dynamic world, is one such problem. While our routing problem is not constrained by problems such as time, the agents in the world require efficient algorithms to swiftly plan routes, to prevent players from encountering enemies which are static. In the real-world, there are multiple problems which require optimization to find feasible or better than feasible solutions. Vehicle routing problems, first defined by Dantzig et al \cite{dantzig1959truck}, is similar to other complex problems such as the Travelling Salesman Problem (TSP) in that an agent is placed at a starting point in a network and, following some constraints, must reach a certain goal. Constraints such as backtracking restrictions or maximum visits to each location, or even additional stops, can be applied. In the survey paper by Pillac et al \cite{pillac2013review} a large number of sub-domains are presented, including the transportation of people, goods and services, where the need for optimization is clear. Additional real world problems for routing can be found in the expanding domain of network routing. Such research areas include \ac{D2D} network routing \cite{shaikh2018routing}, unmanned aerial vehicle networks \cite{arafat2019routing} and disaster relief networking \cite{jahir2019routing}. Each of these research areas have differing constraints, with for example, disaster relief having requirements for redundancy and flexibility that other network optimizations may consider less important.
   
   Similar to routing problems is circuit planning, due to the need to prevent overlapping signals, which is analogous to preventing transport belts from passing materials from one to the other. An interesting outcome from a landmark paper in circuit design optimization using metaheuristics required a user-supplied cost function, placing the onus of constraint enforcement and relaxation on the user \cite{gielen1990analog}. This is similar to how our constraints are provided; before analysis has begun on feasibility. A more recent problem of circuit design and efficiency used \ac{PSO}, an effective algorithm for non-linear exploration of a problem space, to discover efficient pathways for signals \cite{fakhfakh2010analog}. Analog block design is qualitatively evaluated from a perspective of maximizing performance, rather than minimizing cost, though the two may be causally related. 

\section{Problem Description}\label{section:problemdescription}
 
    There are a variety of problems in Factorio which can be automated and optimized through the use of metaheuristics. For this initial work, we study a set of problems from the perspective of direct optimization and machine learning. These methods differ with respect to their goal. Direct optimization searches in the space of solutions directly for a single point which minimizes a problem-specific cost function. Machine learning, on the other hand, optimizes a \textit{prediction machine} which accepts a representation of the problem as input and extrapolates, or \textit{predicts}, the optimal solution point relative to the input state. As such, the goal of direct optimization is to discover a single optimal solution relative to a finite training data-set (Section \ref{subsection:psa}), while the goal of machine learning is to discover a solution that generalizes to unseen data (Sections \ref{subsection:qgp} and \ref{subsection:erl}). Put another way, direct optimization builds \textit{solutions}, while machine learning builds \textit{problem solvers}. The later approach potentially incurs a significantly greater computational cost because a larger number of training examples, and therefore model evaluations, are required to build generalized behaviors.  

\begin{figure}[ht]
  \caption{12 x 12 matrix with obstacles}
  \includegraphics[scale=0.2]{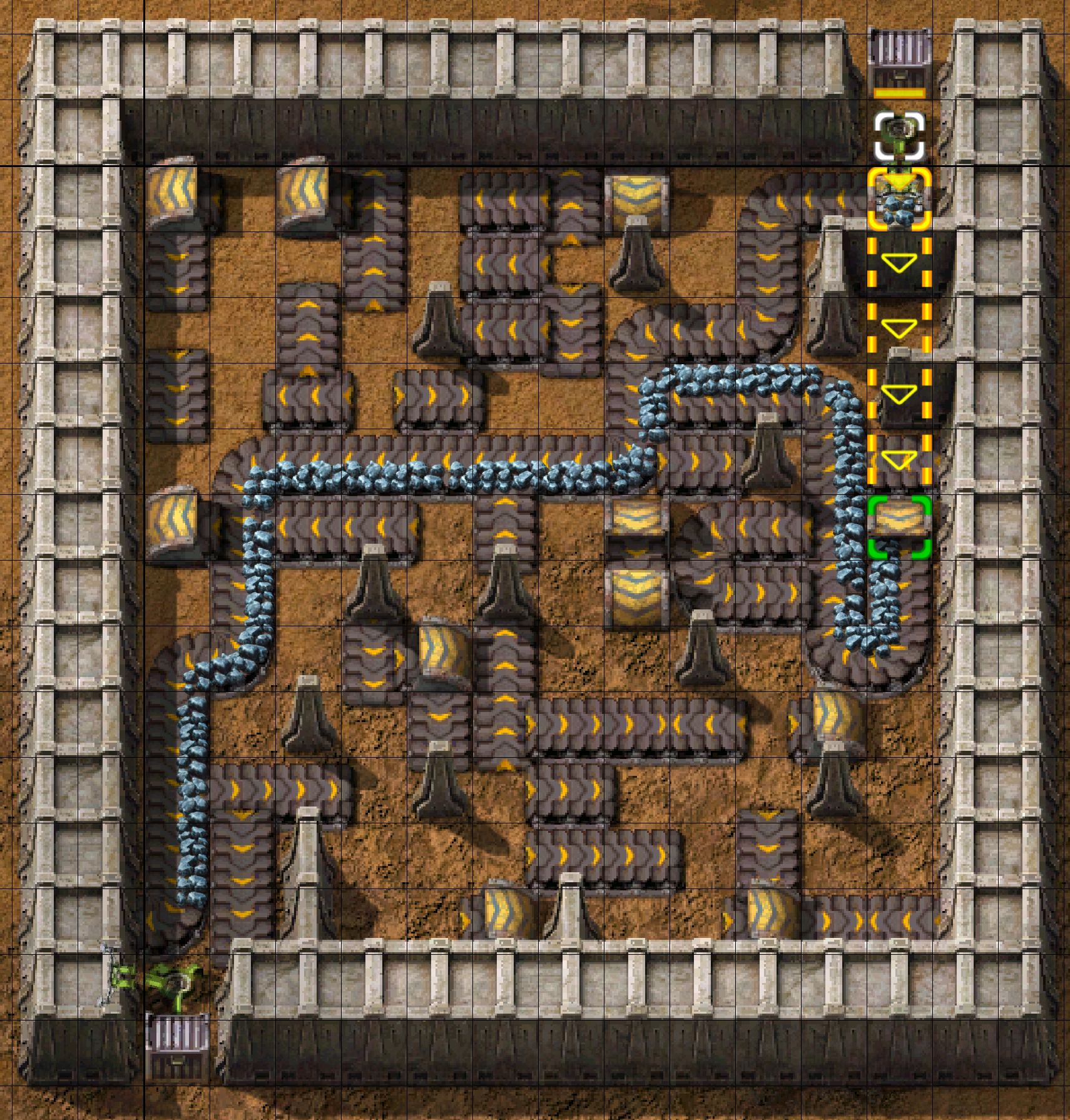}    
  \label{img:screenshot}
\end{figure}

    In this paper, we focus on the logistics belt placement problem. In order to best represent the various aspects of belt placement, we define six problem instances with varying attributes. The problem matrices provide empty cells in sizes of $3 \times 3$, $6 \times 6$ and $12 \times 12$. Outside of these empty matrices are walls, two cells thick, an inserter (for input) with a chest, and a receiver (for output) with a chest. For a visual representation, see the in-game representation in Fig \ref{img:screenshot}, where the input inserter and chest are at the top right of the image, and the output receiver and chest are at the bottom left. Underground belts function by requiring an input and an output belt to be in the same row or column, face the same direction, and have a maximum of 4 squares between the two items. Without this criteria being met, underground belts become nothing more than obstacles.

    As well as the three aforementioned empty grids, we tackle three other problem matrices which are identical in size but have obstacles positioned where logistic items cannot be placed. This both simplifies and complicates the problem, as there are fewer possible solution combinations, with some of them no longer able to transport materials from source to destination.  

\begin{figure}[ht]
  \caption{Integer Encodings}
  \includegraphics[scale=0.5]{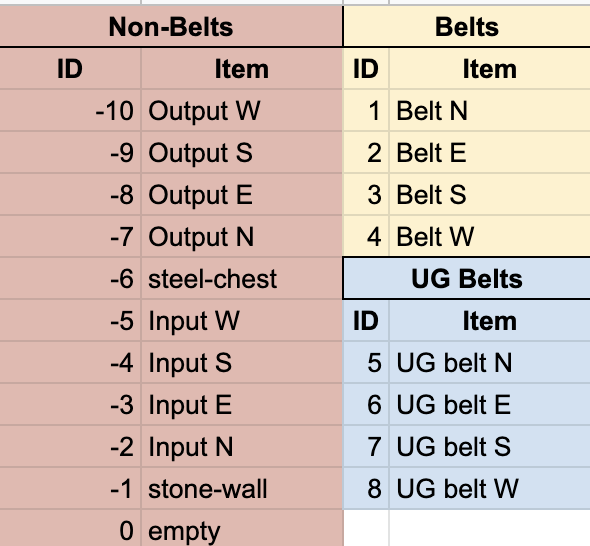}    
  \label{img:matrix}
\end{figure}

    For accessible manipulation by external optimizers, the solutions and problems are encoded as integer matrices. The interface, described in Section \ref{section:implementation}, translates the integers (representations of which are shown in Fig \ref{img:matrix}) and provides commands to Factorio to create the problems and solutions in-game for experimentation. Integers less than zero are not utilized by solutions, but instead are used to represent the problem. 

    The quality of a solution is measured through its fitness value. A fitness of 0 is assigned if a solution is deemed infeasible (having failed one or more hard constraints (HCs), or a value between 0 and 1 otherwise. The hard constraints can be listed as follows:
\begin{itemize}
    \item HC1: A maximum of one logistic object can be placed in a single position at a time or one obstacle. 
    \item HC2: There must be at least one logistic item placed for a solution to be deemed feasible. 
\end{itemize}

\noindent If a solution is feasible, then each \ac{SC} can be evaluated to return a single fitness value from the objective function.
The soft constraints can be listed as follows:
\begin{itemize}
    \item SC1: Maximize the number of item output by the solution.
    \item SC2: Maximize the number of item input into the solution.
\end{itemize}

\section{Mathematical Model}

    This section of the paper concisely defines the problem in a mathematical model, though each algorithm in this paper has a unique implementation. Table \ref{tab:Params} provides the meaning for the mathematical objects in use. 

   \begin{table}
\begin{tabular}{|c|p{6.16cm}|}
    \hline 
    \textbf{Parameters} & \textbf{Description}\\
    \hline
    $M$ & A parameterized matrix of two vectors, $M_x \in M$ containing all $x$ coordinates and $M_y \in M$ containing all $y$ coordinates. Coordinates (position) of an individual position are referred to as $p$, which is a vector with a single $x$ coordinate and $y$ coordinate. \\ 
    \hline
    $d$ & Cardinal direction $d \in \{0, 1, 2, 3\}$ which represent north, east, south and west, respectively. \\
    \hline
    $b_{pd}$ & A belt which can receive 0 or 1 inputs, and 0 or 1 outputs. \\
    \hline
    $u_{pd}$ & An underground belt which can receive 0 or 1 inputs and 0 or 2 outputs. \\
    \hline
    $L$ & An array of two vectors containing all belts and underground belts. \\ & $b \in L_b$, $u \in L_u$ \\ 
    \hline 
    $w_k$ & Weightings for each \ac{SC}. \\ & $k \in \{1, 2\}$. \\ &$w_k \in \mathbb{R}, ~~  k \in \{1,2\}$ \\ & $0 \leq w_k \leq 1, $~~~~$  k \in \{1,2\}$ \\ & $w_1 + w_2 = 1$\\
    \hline
    $\phi_p$ & Obstacle at position $p$. Equal to 1 if there is an obstacle, 0 if not. \\ 
    &$\phi \in \Phi$. \\
    \hline
    $i$ & A parameterised number of items to be input to the solution. \\ & $i > 0$, $i \in \mathbb{N}$ \\ 
    \hline
    $o$ & A parameterised number of items to be output by the solution. \\ & $o > 0$, $o \in \mathbb{N}$ \\ 
    \hline
\end{tabular}
\caption{\label{tab:Params}Parameters}
\end{table}

    Decision variables are the mechanism for directing the solution towards a higher quality and feasible solution. Decision variable \eqref{equation:decVar1} is used for placing a belt at any position $p$ in the provided grid $G$, of any type $c$ and of any direction $d$. A belt can either be placed, $1$, or not, $0$. Similar decision variables are required for underground belts $u$, equation \eqref{equation:decVar2}. 

\begin{flalign} \label{equation:decVar1}
    X_{b_{cpd}} = 0 ~\textrm{or}~ 1 \textrm{ , } &\forall b \in L_b, \\ \nonumber &\forall p_x \in G_x, \forall p_y \in G_y,\\ \nonumber &\forall d \in \{0,1,2,3\} &&
\end{flalign}

\begin{flalign} \label{equation:decVar2}
    X_{u_{cpd}} = 0 ~\textrm{or}~ 1 \textrm{ , } &\forall u \in L_u,\\ \nonumber &\forall p_x \in G_x, \forall p_y \in G_y,\\ \nonumber &\forall d \in \{0,1,2,3\} &&
\end{flalign}

    The objective function is defined as maximizing the sum of each weighted \ac{SC}. 

\begin{flalign}
    \textrm{Max:}~~ (SC1 + SC2)&&
\end{flalign}

\noindent Subject to:
\begin{flalign}
    \textrm{HC1:}~~ &\Bigg[\phi_p +\Bigg(  \mathlarger{\sum}_{b \in L_b} X_{b_{p}}\Bigg) +  \Bigg(\mathlarger{\sum}_{u \in L_u} X_{u_{p}}\Bigg)\Bigg] < 2 ~~ \\ \nonumber &\forall p_x \in M_x, \forall p_y \in M_y && \\ \nonumber &\forall p1_x \in M_x, \forall p1_y \in M_y && \\ \nonumber & \forall \phi \in \Phi &&
\end{flalign}

\begin{flalign}
    \textrm{HC2:}~~ &\Bigg[\Bigg(\mathlarger{\sum}_{b \in L_b} X_{b}\Bigg) + \Bigg(\mathlarger{\sum}_{u \in L_u} X_{u}\Bigg)\Bigg] > 0 ~~ &&
\end{flalign}

\begin{flalign}
    \textrm{SC1:}~~ o * w_1 &&
\end{flalign}

\begin{flalign}
    \textrm{SC2:}~~ i * w_2 &&
\end{flalign}

\section{Implementation}\label{section:implementation}

    The source code of Factorio cannot be modified directly; instead there are three mechanisms where users can provide input to the game. These are 1) through mouse and keyboard control, which is the standard use case for Factorio players, 2) through commands delivered via the in-game console, 3) via `mods' (short for modifications). Direct usage of none of these mechanisms are sufficient for an optimizer to interact with Factorio and evaluate possible solutions to in-game problems. In section \ref{factorioInterface} we introduce \ac{FOI}, an interface which uses \ac{RCON} protocol and modding to connect to the Factorio in-game console and enables optimizers, regardless of their programming language, to easily interact with a running Factorio server.
    
    Modding (the act of writing mods) is often supported by video game companies. With Factorio, the company has provided a modding interface through the scripting language of Lua. While restricted to prevent run-time errors and modification of core elements of the Factorio software, Lua provides structured access to modifying many elements of Factorio. For example, many mods are written to add graphical elements, music or sound effects, additional game-play features such as new vehicles or modifying how existing mechanics function. 
    
    \ac{RCON} \cite{rconsource} is a TCP-based protocol which uses a request-response communication method to enable server conductors to remotely administrate servers. \ac{RCON} establishes a connection by using a simple authentication method which includes sending a password to a server and receiving a response. Upon success, a connection is established and the conductor is able to communicate with the server by sending request packets and receiving responses \cite{rconsource}.
   
    \subsection{Factorio Optimizer Interface}   \label{factorioInterface}
   
        We introduce \ac{FOI}, an open-source cross-platform interface written in Python that uses \ac{RCON}  protocol to enable problem solvers to connect and interact with a running Factorio server console. \ac{FOI} incorporates file system and command-line interface, two programming language-independent methods, to communicate with problem solvers.

\begin{figure}[ht]
      \caption{Diagram of \ac{FOI}}
      \includegraphics[scale=0.60]{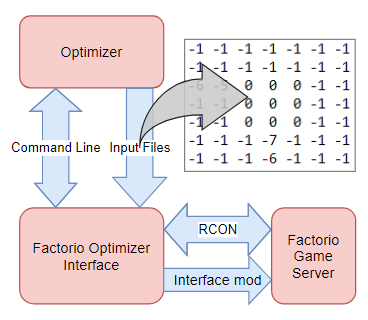}    
      \label{img:DiagramInterface}
\end{figure}

        Figure \ref{img:DiagramInterface} illustrates an optimization system that uses \ac{FOI} to interact with the Factorio server. In each iteration, the optimizer prepares matrices representing the problem and solution(s) (\ac{FOI} is capable of evaluating multiple solution matrices at the same time) and calls the interface program giving it the path to the matrices as input. \ac{FOI} then, uses integer encodings (Figure \ref{img:matrix}) of the matrix cells to create corresponding Lua commands to place entities on the game grid as it is explained in section \ref{section:problemdescription}. The \ac{RCON} protocol enables \ac{FOI} to send these Lua commands to the game server to execute them and to receive the server response. For the logistic belt placement problem, server response will show the number of resources taken from the input chest by the input inserter and the number of resources successfully transferred to the output chest. \ac{FOI} sends the server response back to the optimizer so the fitness of the solution matrix can be evaluated using these information. Some entities in Factorio require certain elements to become enabled. For instance, an inserter needs electricity to transfer resources from/to transport belts. Designing power grids is a cumbersome problem in Factorio. To avoid solving that issue while building problem matrices, \ac{FOI} uses the \textit{interface mod} to remove such constraints and minimizes the amount of entities needed to be placed on the Factorio grid for a problem to be running in-game while remaining visually informative.         
   
        In this paper we present a number of metaheuristics which, through the modding interface provided by Wube software and \ac{RCON}, test solutions to the problem. The selected algorithms represent a variety of heuristic types, including reinforcement learning, genetic programming and classical metaheuristic techniques. Due to the universal inputs allowed by the interface presented in this paper, the metaheuristics are implemented in Java, Python and C++. As described in Section \ref{section:intro}, the complexity of the problem can be expanded to include different speeds of logistic machinery, additional belt-types (such as splitters which can be used to pass objects from one belt to many, many to many, or many to one). As such, the selected algorithms which are not simple path-finding algorithms in order to establish an initial benchmark for this combinatorial optimization problem.

    \subsection{Parallel Simulated Annealing}\label{subsection:psa}
   
        \ac{SA} is a classic search heuristic in the history of metaheuristics. This metaheuristic was initially described in 1953 by Metropolis et al \cite{metropolis1953equation} as an adaptation from the Metropolis-Hastings algorithm. This algorithm was selected partly due to it its popularity, allowing this to be a useful insight into this problem and how it can be solved. Additionally, we selected this algorithm due to its usage in real-world problems, such as mobile network routing \cite{saha2017seer}, vehicle routing \cite{vincent2017simulated}, systems-on-chip design flow \cite{gavrilov2018clustering}.
    
        We utilize a version of \ac{SA} with simultaneous solution comparisons known as \ac{PSA}. This is largely due to the inefficient nature of a serial \ac{SA} within the constraints of this problem, as well as due to the proven ability of similar algorithms such as Differential Evolution to solve larger search spaces, for example by Whalen et al \cite{whalen2020evolving}. In order to facilitate this, while still maintaining the core attributes of \ac{SA}, the algorithm works as can be seen in the Algorithm \ref{pseudocode:SA}.\footnote{The variable lettering in this section is unrelated to other sections of this paper.}
   
        The number of iterations the \ac{PSA} algorithm will perform is calculated as $\ceil[\big]{\log{(\delta / t)} / \log{(cr)}}$. The selection of a neighbor in line \ref{algstep:selectNeighbor} is conducted by modifying a bounded-random number of entities, based on the current number of iterations compared to the total number of iterations to be carried out. This allows the potential number of changes to the solution integer array to be bounded highly at the beginning of run-time. As the temperature $t'$ reaches $k$ the number of changes to select a neighbor will be much lower, with a lower bound of one. 
   
        Fitness is calculated in lines \ref{algstep:fitnessOriginal} and \ref{algstep:fitnessNew} by evaluating each solution in the vectors, and enumerating inputs and outputs each solution allows. 
  
        Following traditional \ac{SA} functionality, solutions are always accepted if fitness is improved, and may be accepted if fitness worsens, depending on run-time and potential fitness loss. The method this decision is described in the function at line \ref{algstep:acceptanceProbability} $n$ times.
   
\begin{algorithm}
    \DontPrintSemicolon
    \caption{Parallel Simulated Annealing}
    \label{pseudocode:SA}
    \KwData{$t$ Initial temperature}
    \KwData{$t'$ Current temperature}
    \KwData{$\delta$ Delta}
    \KwData{$cr$ Cooling Rate}
    \KwData{$n$ Solutions per iteration}
    \KwData{$\vec{S}$ Previous solutions}
    \KwData{$\vec{S'}$ Current solutions}
    \KwData{$\vec{S''}$ Temporary solution vector}
    
    $\vec{S} \leftarrow stochasticallyGenerateNSolutions(n)$\;
    
    \While{$t' > \delta$}{ 
        \For{$i \leftarrow 0;\ i < n;\ i \leftarrow i + 1$}{
            $\vec{S'}_i \leftarrow selectNeighborOf(\vec{S}_i$)\; \label{algstep:selectNeighbor}
        }
        $\vec{F} \leftarrow calculateFitness(\vec{S})$\; \label{algstep:fitnessOriginal}
        $\vec{F'} \leftarrow calculateFitness(\vec{S'})$\; \label{algstep:fitnessNew}
        
        \For{$i \leftarrow 0;\ i < n;\ i \leftarrow i + 1$}{\label{algstep:forloop}
            \If{$\vec{F'}_i >= \vec{F}_i$}{
                $\vec{S''}_i \leftarrow \vec{S'}_i$\;
            }
            \ElseIf{$acceptanceProbability(t',\vec{F'}_i, \vec{F}_i)$}{
                $\vec{S''}_i \leftarrow \vec{S}_i$\;
            }
        }
        $\vec{S} \leftarrow \vec{S''}$
        
    }
    
    \SetKwProg{Fn}{Function}{:}{}
    \SetKwFunction{FMain}{acceptanceProbability}
    \Fn{\FMain{$t'$,$f'$,$f$}}{ \label{algstep:acceptanceProbability}
       $r \leftarrow randomDouble()$\;
       $b \leftarrow 1 - f$\;
       $b' \leftarrow 1 - f'$\;
       
       \If{$r > e^{b' - b} / t'$}{
        \Return{true}\;
        }
        \Return{false}\;
  }
  
\end{algorithm}
   
    \subsection{qGenetic Programming}\label{subsection:qgp}

        \ac{GP} is an evolutionary algorithm which evolves computer programs to solve a problem that can be computationally represented. Initially a population of individuals is created and their fitness is evaluated. The best individual(s) are preserved and a portion are selected based, determined by fitness, to produce the next generation of individuals using evolutionary operators such as crossovers and mutations \cite{Koza1992}.
   
        We present \ac{qGP} for the first time in this paper to solve the logistic belt problem (algorithm \ref{pseudocode:qGP}). This algorithm represents the solutions with a linear genome consisting of a mixture of strings and integers that can be mapped to operators (functions or non-terminals) and operands (terminals). The goal is to interpret each genome to connect as many operators as possible in the order of their appearance with suitable operands to form a competitive model. For instance, a simple summation operator requires two numeric operands to produce a meaningful output. The output of each function then, can be considered as an operand for the next possible operators. \ac{qGP} neglects occurrence of an operator in case the algorithm fails to find fitting operands for it. A sequence of connected operators form a solution or a model. 
        
        In order to interpret a genome \ac{qGP} uses a mapping table to translate each base value to an operator or an operand. There is no limit to the data types that can be used as operands in \ac{qGP}. In fact, it is designed so new data types for operands and/or operators with unique functionalities can be easily added to the optimizer. \ac{qGP} creates a list of all the operands and operators in the order they appear in the genome. Starting from the first operator, if the algorithm can satisfy the operator's operand requirements, then it will use the operator on a set of applicable operands. The operator will then be output to the end of the operands list. If the operator demands cannot be satisfied, \ac{qGP} saves the operator in a different list called \textit{idle operators}. If a new value is added to the operands list, idle operators become eligible for consideration again to check if the new operand will enable them. In \ac{qGP}, unlike many \ac{GP} representations, operators can have multiple outputs or inputs which allows it to build complex model structures. Additionally, outputs of operators can be used multiple times with a single occurrence in the genome. This promotes modularity in \ac{qGP} models while keeping the genome as compact as possible. This is similar to gene duplication in nature but without any overhead in the genome. After interpreting models into combinations of operators and operands (in our case, sequence of calls to \textit{Connect()} and \textit{Place()} functions), \ac{qGP}  saves the matrices generated from each individual to give to \ac{FOI} for fitness evaluation. 
        
        Table \ref{tab:qgpParams} lists the parameters and configurations used in this paper to undertake the logistic belt problem using \ac{qGP}. We use two operators named \textit{Place} and \textit{Connect} in our function set\textit{Place} requires a matrix, a grid position, and an object (a belt or underground belt) as an input. The operator then places the object on the given position of the input matrix and returns the enhanced matrix as an output. The \textit{Connect} operator requires two grid positions and a matrix as input to logistically connect the two positions using transport belts. \textit{Connect} can place multiple objects on the grid simultaneously while \textit{Place} only assigns one object on a grid at a time. We use Tournament Selection \cite{Blickle1995} of size five and elitism \cite{Poli2008} of one individual in forming the next generation of solutions.
    
    \begin{table}
    \begin{tabular}{|c|p{4.5cm}|}
    \hline
    \textbf{Parameter}      & \textbf{Value}                                       \\ \hline
    Population Size         & 30                                                   \\ \hline
    Initial/Max Genome Size & 40/100                                               \\ \hline
    Tournament Size         & 5                                                    \\ \hline
    Elitism         & 1 individual                                                    \\ \hline
    Operators               & Connect, Place                                       \\ \hline
    Operands                & 4 belts and 4 underground belts for each direction, problem matrix, integers (limited to problem matrix size) \\ \hline
    \end{tabular}
    \caption{\label{tab:qgpParams}qGP Parameters}
    \end{table}
   

    \begin{algorithm}
    \DontPrintSemicolon
    \caption{qGenetic Programming}
    \label{pseudocode:qGP}
    \KwData{${g}$ Current generation}
    \KwData{${G}$ Maximum number of generations}
    \KwData{${P}$ Population of solutions}
    ${P} \leftarrow initializePopulation()$\;
    $best \leftarrow 0$\;
    \While{$g < G$}{ 
        \For {$i \leftarrow 0;\ i < size(P);\ i \leftarrow i + 1$}{
            $interpretModel(P[i])$\;
            $createMatrices(P[i])$\;
            $evaluateFitness(P[i])$\;
            \If{P[i].fitness > best}{
                $best \leftarrow P[i].fitness$\;
            }
        }
        $newPop \leftarrow []$\;
        \While{size(newPop) < size(P)-1}{
            $\vec{selected} \leftarrow tournamentSelection(P)$\;
            $newPop \leftarrow newPop + crossover(\vec{selected})$\;
        }
        $newPop \leftarrow mutate(newPop)$\;
        $newPop \leftarrow newPop + best$\;
    }
\end{algorithm}
   
    \subsection{Evolutionary Reinforcement Learning}\label{subsection:erl}
   
        \ac{RL} is a type of machine learning that simulates the way living organisms adapt through trial-and-error interaction with their environment over time\cite{sutton18}. When modeling this interaction in a computer simulation, time is divided into discrete steps. At each timestep, the digital organism, or \textit{agent}, observes its environment, takes an action that changes the state of the environment, and receives a feedback signal that describes the desirability of its current situation. The goal in \ac{RL} is to develop agents that map observations to actions such that the summed feedback, or \textit{reward}, over all timesteps is maximized. Most real-word decision-making problems can be modeled in this way, including game playing. In this work, Factorio provides a rich ecosystem of problem environments, and agents are represented by computer programs. This Section describes how we can use \ac{GP} in combination with \acl{RL} to evolve programs that solve multiple problems in Factorio by interacting with the game. In brief, a population of programs will be stochastically initialized, evaluated in the Factorio environment, selected for quality, and modified through an iterative process inspired by the Darwinian principle of natural selection. This \ac{ERL} process is summarized in Algorithm \ref{pseudocode:erl}.
   
        Interactions with a video game are naturally episodic. The initial state of the environment is set by the game engine and the player interacts with the game until a terminal state is reached (e.g. Game Over) or a time constraint is exceeded. Human players typically need to play multiple episodes in order to gain the breadth of experience required to become a skilled player. Being naturally curious, they may try something new in each episode, leading to a diversity of learning experiences. By contrast, our game-playing programs behave deterministically: given the same initial conditions and a deterministic environment, the programs will produce the same sequence of actions in every episode. Thus, to ensure the programs are exposed to a variety of scenarios, each program is evaluated in multiple episodes with unique environmental conditions (See Algorithm \ref{pseudocode:erl}, lines \ref{tpgstep:genProblems} to \ref{tpgstep:endeval}). In this work, each episode represents a unique belt problem. Then $genProblem$ function in Algorithm \ref{pseudocode:erl} randomly generates problem matrices of a given size, with or with obstacles. The quality, or \textit{fitness}, of a program is measured by its average reward over all episodes it experiences during its lifetime. Thus, programs that perform well in as many belt problems as possible will be favored by selection operators. Programs are evolved using a generational \ac{GA}. In each generation, every program is evaluated in 5 unique episodes, up to a lifetime maximum of 100 episodes per individual (that is, if they survive for 20 generations). After all programs are evaluated, they are sorted by fitness and the worst 50\% are deleted from the population (Algorithm \ref{pseudocode:erl}, line \ref{tpgstep:cull}). New programs are created by uniformly sampling two parents from the surviving individuals and applying crossover and mutation operators. This processes is repeated until the population is back to $P_{size}$(Algorithm \ref{pseudocode:erl}, lines \ref{tpgstep:variation} to \ref{tpgstep:endvariation}). Over time, this combination of GP and RL develops generalized game-playing programs capable of solving multiple unique belt problems in Factorio. The particular \ac{GP} framework we employ for this study is \ac{TPG}. No problem-specific functions were used. A more detailed account of \ac{TPG} in \ac{RL} and video game playing can be found in \cite{kelly18a}.
   
        \subsubsection{RL Interface and Evaluation}
   
            The observation space in our formulation of Factorio belt problems takes the form of an integer matrix (See Figures \ref{img:matrix} and \ref{img:screenshot}). However, the \ac{RL} agents in this work operate from a partial world-view. At any timestep, the agent is situated in one cell of the problem matrix and its observation space, $\vec{s}(t)$, consists of the 3x3 matrix of cells centered at the currently occupied location, for a total of 9 integer state variables in $\vec{s}(t)$. The agent is initialized in the center of the matrix at the beginning of each episode. The location of inputs and outputs will never be visible from this initial state, and the agent must therefore explore its surroundings and possibly retrace its steps when deciding on a path for transport belts\footnote{The TPG variant employed in this work includes a temporal mechanism \cite{kelly2020} which is mostly likely useful in the partially-observable state space, but an ablation study to confirm the significance of memory is left to future work.}. In each timestep, the agent can choose to move one step in any cardinal direction, or place one of 4 types of transport belt at the current location. Thus, thus total action space is $\mathcal{A}={N,E,S,W,1,2,3,4}$, corresponding to one-step movement in any direction or the placement of one type of transport belt at the agent's current location. Transport belts can be placed in any cell to overwrite the previous cell contents, except where this would overwrite a component of the initial problem (i.e. walls, chests, or inserters). Naturally, the agent's location is limited to the bounding box defined by the current matrix size (3x3 or 6x6 in this work). Each episode lasts for 20 timesteps. When this time constraint is reached, the final matrix state is submitted to Factorio for evaluation.
  
\begin{algorithm}
    \DontPrintSemicolon
    \caption{Evolutionary Reinforcement Learning}
    \label{pseudocode:erl}
    \KwData{${g}$ Current generation}
    \KwData{${G}$ Total number of generations}
    \KwData{${P}$ Population of programs}
    \KwData{$P_{size}$ Program population size}
    \KwData{${M}$ Set of problem matrices}
    $P \leftarrow initializePopulation()$\;
    \While{$g < G$}{ 
        $M\leftarrow generateProblems$\;\label{tpgstep:genProblems}
        \For {$i \leftarrow 0;\ i < size(P);\ i \leftarrow i + 1$}{
            \For {$j \leftarrow 0;\ j < size(M);\ j \leftarrow j + 1$}{
                $evaluateFitnessRL(P[i],M[j])$\;\label{tpgstep:episode}
            }
        }\label{tpgstep:endeval}
        $P \leftarrow deleteWorst(P)$\;\label{tpgstep:cull}
        \While{$size(P) < P_{size})$}{\label{tpgstep:variation}
            \If{$randomDouble() < crossoverProbability$}{
                $parent1 \leftarrow sampleUniform(P)$\;
                $parent2 \leftarrow sampleUniform(P)$\;
                $child \leftarrow crossover(parent1, parent2)$\;
            }
            \Else{
                $child \leftarrow sampleUniform(P)$\;
            }
            $P \leftarrow P + mutate(child)$\;
        }\label{tpgstep:endvariation}
    }
\end{algorithm}
   
\section{Results}\label{section:results} 
   
    We utilize different metaheuristics presented in section \ref{section:implementation} to investigate the performance of each algorithm in this novel problem. Testing was conducted on mixed hardware, but each heuristic was given approximately equal maximum computational time by limiting number of iterations. While the authors recognize that an iteration in one heuristic is not equivalent to an iteration in a differing heuristic, the purpose is to investigate varying algorithm types rather than compare similar heuristics and report the best.
   

  \begin{figure}[]
	\centering
	\subfigure[3x3 Problem]{\includegraphics[scale=0.26]{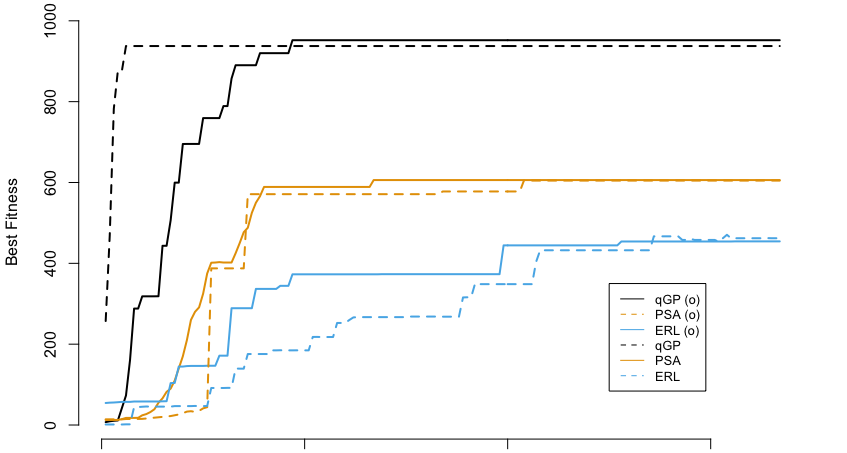}}\newline
	\subfigure[6x6 Problem]{\includegraphics[scale=0.26]{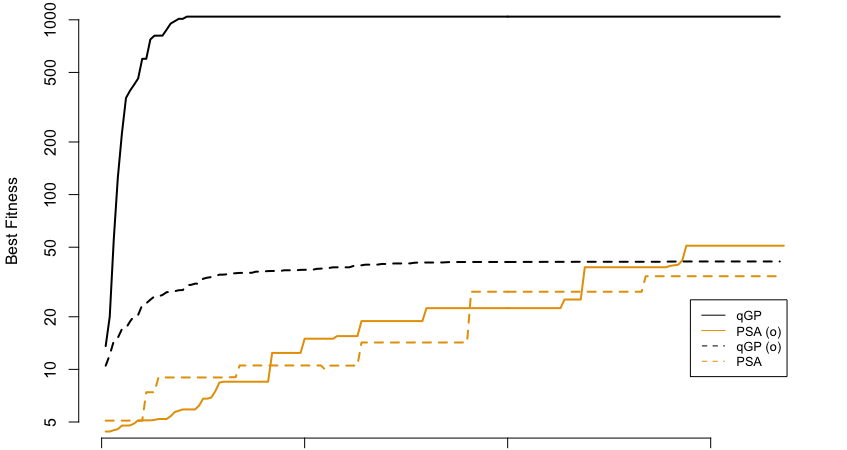}}\newline
	\subfigure[12x12 Problem]{\includegraphics[scale=0.26]{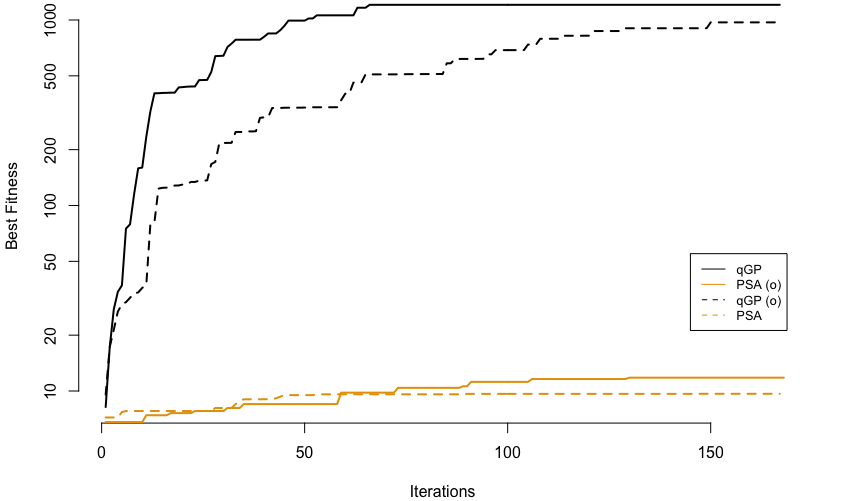}}\newline
	\caption{Average Best-Fitnesses by Iterations. (y-axis of 6x6 and 12x12 plots are log scale)}
	\label{img:trainingCurves}
\end{figure}

        Figure \ref{img:trainingCurves} plots the average best fitness of the best solution discovered by each algorithm over 167  iterations for each problem matrix. \ac{qGP} provides the highest scoring solutions in all cases but one. This is likely a result of the powerful task-specific operators included in the \ac{qGP} function set (See \textit{Connect} Section \ref{subsection:qgp} Table \ref{tab:qgpParams}). However, in the 6x6 problem the obstacles caused the task-specific operators to fail to find a high quality solution. \ac{PSA} successfully finds solutions for the 3x3 problems but cannot find solutions for the 6x6 nor the 12x12 problems. \ac{PSA} has similar performances with and without obstacles, whereas \ac{qGP} shows distinctly different attributes when solving obstacles compared to no obstacles for each of the three problems. In the 3x3 problem (and the 6x6 and 12x12 for \ac{qGP}), the initial steep climb in fitness is due to solutions being created which allow transportation of objects from the input without them being delivered to the output. Additional success once the two points are connected can be due to more efficient pathways allowing more objects to be delivered before allowed run-time completes. Fitness in the solutions such as those produced by \ac{PSA} in the 6x6 problem, for both with obstacles and without, is continuing to climb as the distance covered by additional belts is increased, however they fail to reach the output. Due to the high number of possible changes that can be made (9 integers to select from in each cell, is $9!^{12*12}$), \ac{SA} fails to compete with \ac{qGP}. 
        
\begin{figure}[]
	\centering
	\subfigure[3x3 without obstacles]{\includegraphics[height=3.75cm]{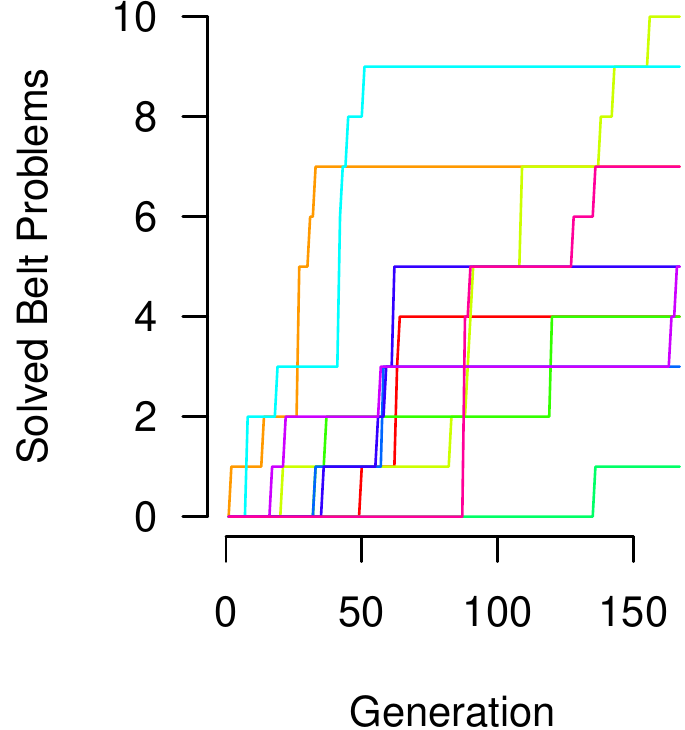}}
	\hspace{0.5cm}
	\subfigure[3x3 with obstacles]{\includegraphics[height=3.75cm]{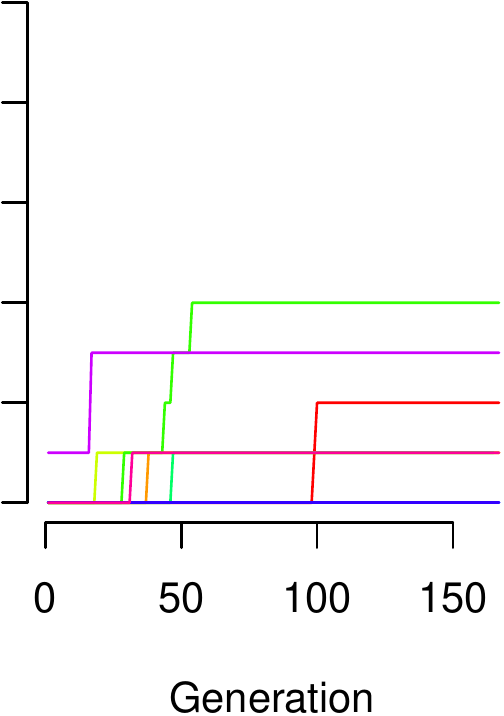}\label{fig:sunspots-st}}
	\hspace{0.2cm}
	\caption{Number of unique belt problems solved by the best \ac{ERL} generalist agent at each generation. A problem is considered solved if the agent manages to connect the input and output with transport belts within 20 timesteps.} 
	\label{fig:erl-solved}
\end{figure}

        Due to the computational cost of evaluations in \ac{ERL}, only the 3x3 matrix problems are considered. \ac{ERL} on average finds the highest fitness solution at a distinctly earlier rate than without obstacles. However, the strength of this approach is found in the generalization ability of the evolved agents, and Figure \ref{img:trainingCurves} simply plots the average best fitness achieved for a \textit{single} problem instance. The generalization ability of \ac{ERL} agents is demonstrated in Figure \ref{fig:erl-solved}, which plots the number of unique belt problems solved by the single best \ac{ERL} agent at each generation (each line represents 1 of 10 independent runs). The problems without obstacles are clearly easier, and the single best agent generalizes to 10 unique problem configurations. When obstacles are introduced, the best agent can only solve 4 unique problems.
        
\section{Future Work}\label{section:futureWork}

    This paper represents an initial exploration into automating and finding high quality solutions to in-game problems through an external interface. Our future work will include a deeper analysis into the logistic belt problem, expanding into the use of other logistic items including splitters, varying speeds of logistics and updating our fitness functions to include considerations for weighted constraints such as cost of materials, size of solution, speed and throughput of transportation of materials. 

    Other problems in Factorio were considered at the beginning of our research. One such problem is the electricity pole placement problem, which is particularly interesting due to the relationship this problem has with placement of machinery (as each item requires an electrical supply). This may be an interesting problem to consider co-evolution techniques. Another example of potential future work includes decision-making problems which consider resource management, power management and pollution generation, and attempts to control automated systems (such as the problem described in this paper) to minimize or maximize certain objectives.
    
    Each of the algorithms in this paper is tested upon the logistic belt problem. Additional extensions to these algorithms could enhance their ability to cope with large search spaces, such as path-finding hybridization. An important feature of \ac{qGP} is the ability to produce multiple outputs with a single operator. Additional complex model structures may provide improvements to tackling the logistic belt problem.

\section{Conclusions}\label{section:conclusion}

    The video game Factorio constitutes an untapped suite of research problems, and this paper provides an initial step into exploring them. In this study we present an introduction to problems which can be found within Factorio and define one of these problems: the logistic belt optimization problem. We provide three algorithms to attempt to solve the problem and compare their efficacy. A mathematical model of the logistic belt optimization problem is presented. This work also provides an interface with Factorio to allow external optimization algorithms to connect to Factorio, allowing future researchers to conduct experiments without experience with the in-game scripting language, Lua. Our results show a variety of algorithm types and the benefits of exploratory algorithms such as Parallel Simulated Annealing, the strengths of domain-focused genetic programming and the flexibility of general purpose metaheuristics, specifically evolutionary reinforcement learning. 

\section{Acknowledgments} 

This work was supported in part by the National Institute of Food and Agriculture (AFRI Project No. 2019-67015-29323), the BEACON Center for the Study of Evolution in Action and  through computational resources and services provided by the Institute for Cyber-Enabled Research at Michigan State University.

\bibliographystyle{splncs04}
\bibliography{bibl}

\end{document}